\documentclass[sigconf]{acmart}
\usepackage{balance}
\usepackage{subfigure}
\AtBeginDocument{%
  \providecommand\BibTeX{{%
    \normalfont B\kern-0.5em{\scshape i\kern-0.25em b}\kern-0.8em\TeX}}}
    
\copyrightyear{2020}
\acmYear{2020}
\setcopyright{acmcopyright}\acmConference[ICMR '20]{Proceedings of the 2020 International Conference on Multimedia Retrieval}{June 8--11, 2020}{Dublin, Ireland}
\acmBooktitle{Proceedings of the 2020 International Conference on Multimedia Retrieval (ICMR '20), June 8--11, 2020, Dublin, Ireland}
\acmPrice{15.00}
\acmDOI{10.1145/3372278.3390694}
\acmISBN{978-1-4503-7087-5/20/06}
\begin{document}
\fancyhead{}

\title{PredNet and Predictive Coding: A Critical Review}

\author{Roshan Prakash Rane}
\authornotemark[1]
\email{rane@uni-potsdam.de}
\affiliation{%
  \institution{University of Potsdam, Potsdam, Germany}}
\author{Edit Sz\"ugyi}
\authornotemark[1]
\email{szuegyi@uni-potsdam.de}
\affiliation{%
  \institution{University of Potsdam, Potsdam, Germany}}
\author{Vageesh Saxena}
\email{saxena@uni-potsdam.de}
\authornote{All authors contributed equally to this research.}
\affiliation{%
  \institution{University of Potsdam, Potsdam, Germany}
 % \streetaddress{P.O. Box 1212}
}

\author{Andr{\'e} Ofner}
\email{andre.ofner@ovgu.de}
\affiliation{%
  \institution{Otto von Guericke University, Magdeburg, Germany}}
\author{Sebastian Stober}
\email{stober@ovgu.de} 
\affiliation{%
  \institution{Otto von Guericke University, Magdeburg, Germany}}


\begin{abstract}
PredNet, a deep predictive coding network developed by Lotter et al., combines a biologically inspired architecture based on the propagation of prediction error with self-supervised representation learning in video. While the architecture has drawn a lot of attention and various extensions of the model exist, there is a lack of a critical analysis. We fill in the gap by evaluating PredNet both as an implementation of the predictive coding theory and as a self-supervised video prediction model using a challenging video action classification dataset. We design an extended model to test if conditioning future frame predictions on the action class of the video improves the model performance. We show that PredNet does not yet completely follow the principles of predictive coding. The proposed top-down conditioning leads to a performance gain on synthetic data, but does not scale up to the more complex real-world action classification dataset. Our analysis is aimed at guiding future research on similar architectures based on the predictive coding theory.
\end{abstract}

\begin{CCSXML}
<ccs2012>
   <concept>
       <concept_id>10010147.10010257</concept_id>
       <concept_desc>Computing methodologies~Machine learning</concept_desc>
       <concept_significance>300</concept_significance>
       </concept>
   <concept>
       <concept_id>10010147.10010257.10010282.10011305</concept_id>
       <concept_desc>Computing methodologies~Semi-supervised learning settings</concept_desc>
       <concept_significance>300</concept_significance>
       </concept>
   <concept>
       <concept_id>10010147.10010178.10010224.10010240.10010244</concept_id>
       <concept_desc>Computing methodologies~Hierarchical representations</concept_desc>
       <concept_significance>300</concept_significance>
       </concept>
   <concept>
       <concept_id>10010147.10010178.10010224.10010245</concept_id>
       <concept_desc>Computing methodologies~Computer vision problems</concept_desc>
       <concept_significance>300</concept_significance>
       </concept>
 </ccs2012>
\end{CCSXML}

\ccsdesc[300]{Computing methodologies~Machine learning}
\ccsdesc[300]{Computing methodologies~Semi-supervised learning settings}
\ccsdesc[300]{Computing methodologies~Hierarchical representations}
\ccsdesc[300]{Computing methodologies~Computer vision problems}
\keywords{deep learning, convolutional neural networks, video classification, video prediction, semi-supervised, predictive coding}


\settopmatter{printfolios=true}
\maketitle

\section{Introduction}

Learning a model of the visual world is a crucial prerequisite to reliably perform computer vision tasks like object detection and semantic segmentation. As illustrated by \cite{DBLP:journals/corr/abs-1902-06162}, self-supervised learning allows to extract this complex structure of the real-world without a need for expensive labeled data. Videos contain information about how scenes evolve in time and therefore predicting the future frames of a video is one popular method \cite{finn16}\cite{mathieu15}\cite{vondrick18}\cite{wang16} \cite{whang18} of extracting this structure in a self-supervised manner. Previous research \cite{lotter16}\cite{mathieu15}\cite{srivastava15} has hypothesized that to accurately predict how the visual world changes, a model should learn about the object structure and the possible transformations objects can undergo. Among the various video prediction models, PredNet by Lotter et al.\cite{lotter16} achieves high video prediction accuracy with the additional benefit of using a biologically plausible architecture.

The PredNet architecture is inspired by the predictive coding theory from the neuroscience literature  
\cite{Friston2009}\cite{rao99}\cite{spratling2008} and attempts to implement it with deep neural networks. Predictive coding is a promising self-supervised learning technique and has shown to replicate some of the neuronal behavior seen in the mammalian visual cortex. It posits that the brain is continually making predictions of incoming sensory stimuli and uses the deviations from these predictions as a learning signal. It describes hierarchical networks consisting of top-down connections that carry predictions from higher to lower levels and bottom-up connections that carry sensory evidence from lower to higher levels at each layer. The error in prediction is propagated upwards, eventually leading to better predictions in the future. 

In this paper, we first evaluate PredNet's performance on video prediction by testing it on a demanding dataset. Then we examine its capability to learn latent features that are useful for downstream tasks. Specifically, our contributions in this paper are two-fold:
\begin{enumerate}

    \item Using visualization techniques and experiments, we review PredNet as an emulation of the predictive coding framework and as a video prediction model.
     
    \item We test the features extracted by PredNet by training it in a semi-supervised setup to perform video action classification. We also evaluate if the conditioning of top-down predictions on action classes of the video and vice versa improves the model's accuracy. We further test this hypothesis on a simple synthetic dataset by conditioning the predictions of PredNet with informative top-down class labels.
     
\end{enumerate}

The paper is organized as follows: Section \ref{rel_work} reviews predictive coding and its implementations. Section \ref{sec:exp_setup} describes our experiment setup, namely our data, the architecture and the evaluation metrics used. Section \ref{sec:probe} is dedicated to the first phase of our experiments, listing our observations while testing PredNet. Section \ref{sec:lab_clas} gives details on the second phase, the implementation and evaluation of PredNet+, our proposed extension of the architecture, and in Section \ref{sec:conc} we conclude the paper and list possible directions for future research.  

\section{Related work}\label{rel_work}
Rao and Ballard \cite{rao99} replicated the `extra-classical' receptive field effects observed in the early stages of cat and monkey visual cortex with an artificial predictive coding network. These observable effects are a direct result of the brain trying to efficiently encode sensory data using prediction. This was accompanied by a rich body of work in neuroscience \cite{Clark2013}\cite{Emberson2015}\cite{Friston2009}\cite{spratling2008}\cite{summerfield2008} and computational modelling \cite{Chalasani2013}\cite{lotter16}\cite{Song2019} that explored different interpretations and implementations of the basic idea, not only in the visual domain but also in various sensory-motor domains. 

Following the success of deep learning in the last decade, many researchers have attempted to implement the predictive coding model using deep learning \cite{han18} \cite{lotter16} \cite{CPC2018} \cite{wen18}. Wen et al. \cite{wen18} use predictive coding on static images to learn optimal feature vectors at each layer for object recognition. Han et al. \cite{han18} build on this to develop a bidirectional and dynamic neural network with local recurrent processing. A. Oord et al. \cite{CPC2018} perform predictive coding in a latent space and use a probabilistic contrastive loss to learn useful representations. Lotter et. al. \cite{lotter16} design a video prediction network using the principles of predictive coding. We chose to evaluate Lotter et al.'s PredNet architecture for two main reasons: First, it is designed to be structurally close to the biological predictive coding model with its hierarchical structure, bottom-up error propagation, and top-down predictions. Second, PredNet achieves accuracy on-par with state-of-the-art video prediction models and is a popular baseline used across different spatio-temporal prediction tasks.

Zhong et al. \cite{zhong18} extend PredNet into AFA-PredNet within the robotics domain. They integrate the motor actions of a robot as an additional signal to condition the top-down generative process. Following this, they design MTA-PredNet \cite{zhong18b} that has different temporal scales at different layers in the hierarchy. They developed MTA-PredNet to compensate for PredNet's inability to perform reliable long-term predictions which is a necessity in robotics for planning. Furthermore, researchers have tried to improve PredNet by adding skip-connections alongside error propagation \cite{Sato2018}, reducing the number of parameters by using fewer gates in the top-down ConvLSTM units \cite{Elsayed2019} and also using inception-type units within each PredNet layer \cite{Hosseini2019}. Sato et al.\cite{Sato2018} evaluate PredNet on weather precipitation dataset and Watanabe et al. \cite{Watanabe2018} test PredNet's response to visual illusions to examine whether predictive coding models respond to visual illusions just as humans do. However, none of the above work critically evaluates PredNet as an implementation of predictive coding and as a reliable self-supervised pretraining method. Our work aims to provide this in the form of a critical review of PredNet for future works that intend to use the architecture or design architectures inspired by it.

\section{Experiment setup} \label{sec:exp_setup} 

\subsection{Dataset}
Most existing large scale video classification datasets have coarse-grained labels\cite{heilbron15}\cite{karpathy14}\cite{kuehne11}. This means that the models are trained on a relatively easy task and the label can be detected even from isolated frames, e.g. the `soccer' label can be inferred from a green field. To overcome this issue and force models to learn better representations, the Something-something dataset \cite{goyal17} was introduced. This dataset contains 220,000 videos with 174 fine-grained action labels. For instance, `putting something on a table', `pretending to put something on a table', and `putting something on a slanted surface so it slides down' are three different label classes in the dataset. Mahdisoltani et al. \cite{farzaneh18} provide evidence for the hypothesis that task granularity is strongly correlated with the quality and generalizability of learned features. As for the nature of the data, being crowd-sourced, it includes noise much resembling the real world: thousands of different objects, variations of lighting conditions, background patterns and camera motion.

\subsection{PredNet architecture}
\begin{figure}[t]
\begin{center}
   \includegraphics[width=0.8\linewidth]{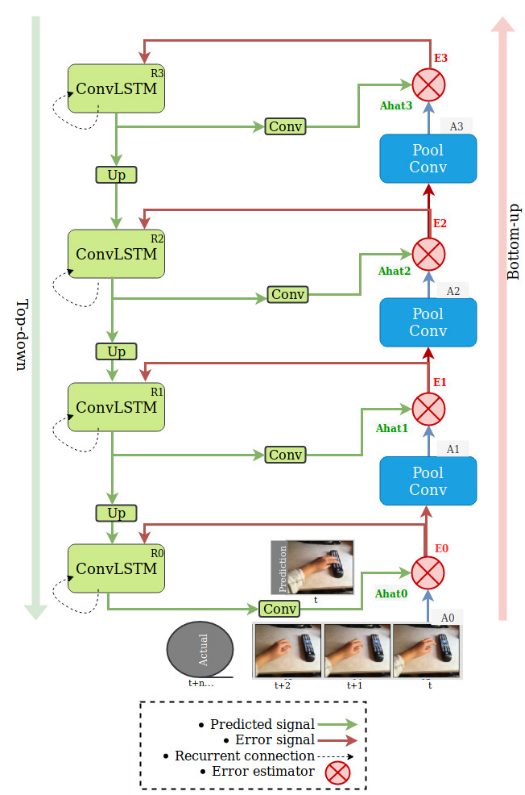}
\end{center}
   \caption{The original PredNet architecture by Lotter et al. \cite{lotter16}.}
\label{fig:prednet}
\end{figure}

The PredNet architecture is shown in Figure \ref{fig:prednet} \cite{lotter16}. The network is composed of stacked hierarchical layers, each of which attempts to make local predictions of its input. The difference between the actual input and this prediction is then passed up the hierarchy to the next layer. Information flows in three ways through the network: (1) the error signal flows in the bottom-up direction as marked by the red arrows on the right, (2) the prediction signal flows in the top-down direction as shown by the green arrow on the left, and (3) the local error signal and prediction estimation signal flow within each layer. 
Every layer consists of four units: an input convolution unit ($A_{i}$), a recurrent representation unit ($R_{i}$) followed by a prediction unit ($Ahat_{i}$) and an error calculation unit ($E_{i}$) as labelled in Figure \ref{fig:prednet}. The representation unit is made of a ConvLSTM \cite{shi15} layer that estimates what the input will be on the next time step. This input is then fed into the prediction unit that generates the prediction $Ahat_{i}$. The error units calculate the difference between the prediction and the input which is fed as input to the next layer. The representation unit receives a copy of the error signal (red arrow) along with the up-sampled input from the representation unit of the higher-level (green arrow), which it uses along with its recurrent memory to perform future predictions.
%%%%%%%%%%%%%%%%%%%%%%%%%%%%%%%%%%%%%%%%%%%%%%%%%%%%

\subsection{Evaluation metrics} \label{metrics}
Defining a good evaluation metric for the quality of image predictions is a challenging task by itself \cite{Kanjar}\cite{mathieu15}. There is no universally accepted measurement of image quality and consequently for image similarity. For the video prediction task, we employ the two commonly used metrics in literature: Peak Signal Noise Ratio (PSNR) \cite{mathieu15} and the Structural Similarity Index Measure (SSIM) \cite{wang04}. Like Mathieu et al.~\cite{mathieu15}, we calculate PSNR and SSIM only for the frames which have movement with respect to the previous frame and call them `PSNR movement' and `SSIM movement' respectively. In our case this is crucial as action videos often contain very few frames with movement and a metric should not reward a model for simply predicting a still frame. We use a third metric called `conditioned SSIM', which is calculated as given in Equation \ref{eq:ssim}. This metric quantifies how different the predictions are from the previous frame and therefore measures how `risky' the predictions of our model are in comparison to simply performing a `last-frame-copy'.
\begin{multline}
SSIM_{cond} = (SSIM_{max} - SSIM(actual_{t-1}, pred_{t})) \\
* SSIM(actual_{t}, pred_{t}) 
\label{eq:ssim} 
\end{multline}

\section{Probing PredNet} \label{sec:probe}

\begin{figure*}
  \includegraphics[width=\textwidth]{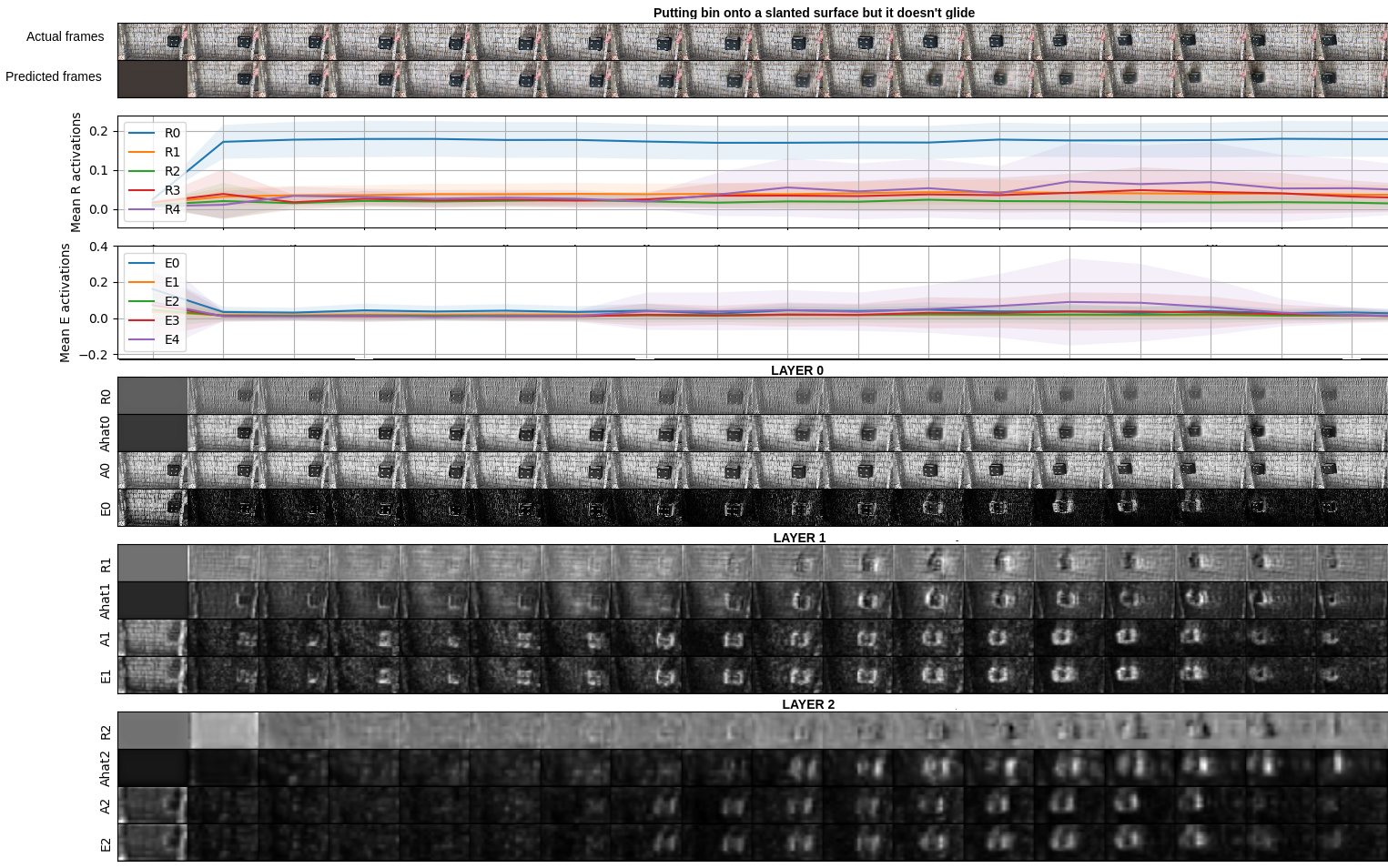}
  \caption{Visualization of evaluated time-specific model states during next frame prediction. Each column corresponds to a single time step, while rows resemble the computed states in each layer.}
  \label{fig:sliding_vis}
\end{figure*}

In the first phase, we review PredNet by evaluating its performance on the Something-something dataset and visualize different states of the architecture. For our experiments we use 10 different hyperparameters settings with a different number of layers, channels per layer, input image size and frames-per-second (FPS) of the video. These settings are listed in Table \ref{tab:models}. Along with the predicted frame, we visualize the different states of PredNet at each layer by averaging the activation of all channels in a layer, similar to Han et al. \cite{han18}. We also plot the mean of the error signals $E_{i}$ and representations $R_{i}$ of every layer to visualize how they evolve over the span of the video. A sample video with visualizations is shown in Figure \ref{fig:sliding_vis}. In the following section, we dedicate one paragraph to each of our findings, and Figure \ref{fig:sliding_vis} and Table \ref{tab:models} provide further details.

\begin{figure}[t]
 \centering
  \includegraphics[width= 1.0\linewidth, height=1cm, keepaspectratio]{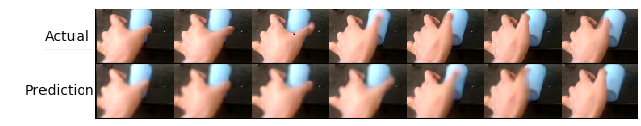}
  \caption{Example of a video with multiple possible future states.}
  \label{fig:rolling}
\end{figure}

\begin{figure}[h!]
\begin{center}
   \includegraphics[width=1.0\linewidth, height=1cm, keepaspectratio]{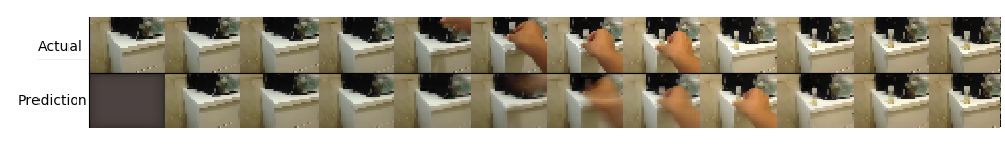}
\end{center}
   \caption{Example of a low FPS video and the predictions made by PredNet.}
\label{fig:fps3}
\end{figure}

\begin{figure}[t]
\begin{center}
   \includegraphics[width=4cm, height=7cm]{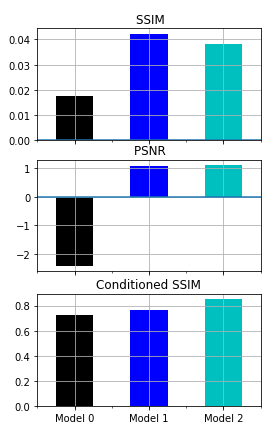}
\end{center}
   \caption{Comparison of model performance with respect to the employed frames-per-second rate (FPS). SSIM and PRNS scores show the model's improvement on a last-frame-copy baseline model.}
\label{fig:fps_comp}
\end{figure}

\begin{table}[h!]
  \begin{center}
    \begin{tabular}{|c|c|c|c|c|c|c|c|c} 
      \hline 
      \textbf{}&\textbf{Frame}&\textbf{}& \textbf{Image } & \textbf{Number} & \textbf{}\\
      \textbf{Model}&\textbf{rate}&\textbf{Layers}& \textbf{size} & \textbf{of} & \textbf{Loss}\\
      &  {(FPS)} &  & {(pixel)} & {\textbf{param.}} & {}\\
      \hline
     0 & \textbf{3} & 4 & 48 X 56 & 6.9 & $L_{0}$ \\
		1 & \textbf{6} & 4 & 48 X 56 & 6.9 & $L_{0}$\\
		2 & \textbf{12} & 4 & 48 X 56 & 6.9 & $L_{0}$\\ \hline
		3 & 12 & \textbf{4} & 32 X 48 & 6.9 & $L_{0}$\\
		4 & 12 & \textbf{5} & 48 X 80 & 5.3 & $L_{0}$\\
		5 & 12 & \textbf{6} & 64 X 96 & 5.8 & $L_{0}$\\
		6 & 12 & \textbf{7} & 128 X 192 & 6.2 & $L_{0}$\\ \hline 
		7 & 12 & 6 & \textbf{96 X 160} & \textbf{7.2} & $L_{0}$\\ \hline	
		8 & 12 & 5 & 48 X 80 & 5.3 &\textbf{$L_{all}$}\\
		9 & 12 & 6 & 64 X 96 & 5.8 &\textbf{$L_{all}$}\\ \hline
    \end{tabular}
    \vspace*{5mm}
    \caption{ Results obtained from different frame rates, number of layers and model parameters (in millions) in the model. Listed are the image size and whether the model was trained with $L_{0}$ loss or $L_{all}$ loss. Similar models are grouped with horizontal lines and the column that varies is marked in bold.}
     \label{tab:models}
  \end{center}
\end{table}

\subsection{Observations}\label{sec:obs}
Comparing the frame predictions with input frames in  Figure \ref{fig:sliding_vis}, Figure \ref{fig:rolling} and Figure \ref{fig:fps3}, we can summarize the working dynamics of PredNet on the action classification dataset as follows: The model performs previous-frame-copy if there are no cues for motion in the previous two frames. If there is a cue for motion and if the direction of the motion is continuous and the motion is smooth, it interpolates the object in the direction of the motion. Otherwise, it blurs the region containing the object of motion to keep the L2 loss minimal by virtue of regression-to-the-mean. The blurring is a result of PredNet's inability to learn multi-modal predictions in the sense that it learns to perform one ideal prediction.  It is a typical characteristic of action-based video sequences that there are multiple possible future states. For instance, as seen in Figure \ref{fig:rolling}, the thumb can move up or move down or not move at all in the next frame. The blurring behaviour of PredNet is further characterized by the experiment we conducted with different sharpness measures. The predictions by the model are always less sharp than the actual videos.

\textbf{PredNet learns relevant features only when trained on videos with continuous motion.} The authors \cite{lotter16} designed and tested PredNet on videos with continuous motion, such as the KITTI dataset \cite{geiger13} and their synthetic `Rotating Faces' dataset. This is in stark contrast with our action dataset which can have a lot of still frames, see e.g. Figure \ref{fig:fps3}. In this scenario, PredNet resolves to mere last-frame copying, as it is statistically beneficial to do just that. If the model is not motivated enough to learn the dynamics of how objects move and scenes evolve then the features it learns would not be useful for downstream tasks as hypothesized by Lotter et al. \cite{lotter16}.

\textbf{PredNet's learning ability is sensitive to the frames-per-second (FPS) rate.} When we compare the performance of models trained on videos with FPS rates 3, 6 and 12 in Figure \ref{fig:fps_comp}, we can see that the performance varies greatly. In this and all following figures, we show the improvement in the given evaluation metric, meaning the difference in the value of the evaluation metric between the model being evaluated and a baseline model performing last-frame copying, i.e. how much improvement the given PredNet model shows to simple last-frame copying. Manual inspection of the predictions further confirms the large difference in prediction quality. At very high FPS there is minimal motion between two consecutive frames while at low FPS rates there is abrupt movement between frames which is challenging to predict. In both of these scenarios, the model completely resorts to last-frame-copy. Therefore, the FPS of the video is one of the most important hyperparameters of PredNet.

\begin{figure}[t]
\begin{center}
   \includegraphics[width=8cm, height=7cm]{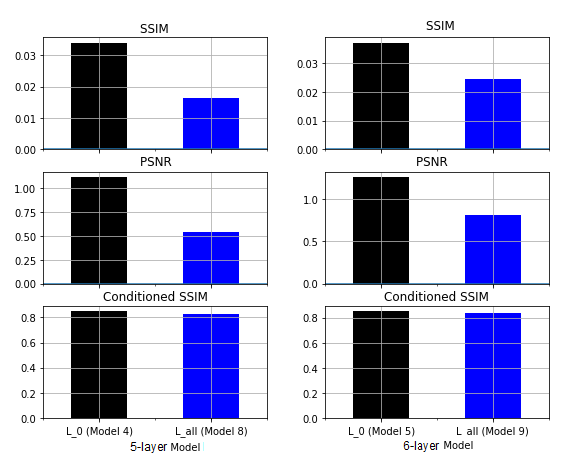}
\end{center}
   \caption{Influence of $L_{0}$ and $L_{all}$ loss on model performance. SSIM and PRNS scores show the model's improvement on a last-frame-copy baseline model.} 
\label{fig:losses_comp}
\end{figure}

\begin{figure*}
  \includegraphics[width=1.0\textwidth]{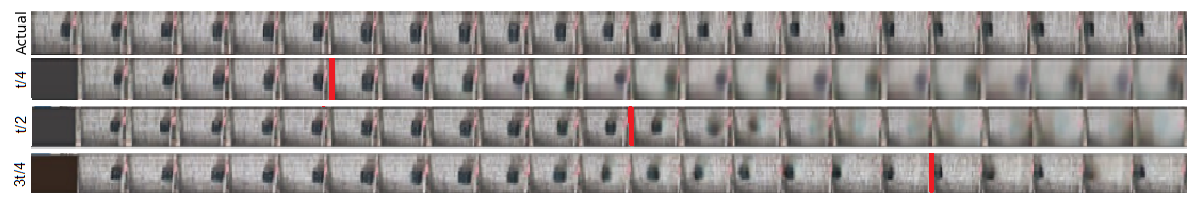}
  \caption{Extrapolation results for Model 7 extrapolated at different time steps. The red mark shows the start of the extrapolation.}
  \label{fig:t+n_extrap_different_timestep}
\end{figure*}

\begin{figure}[t]
\begin{center}
\includegraphics[width=4cm, height=7cm]{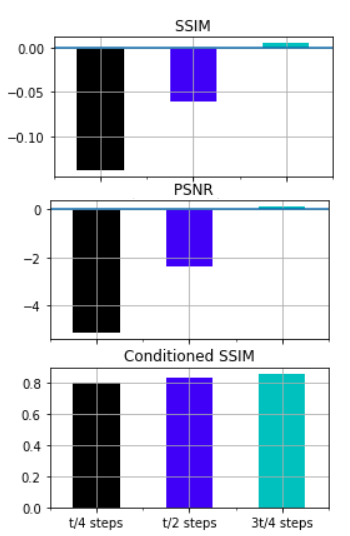}
\end{center}
\caption{Comparison of the metrics when starting extrapolation at different stages in the video. $t$ denotes the total number of frames in the video. SSIM and PRNS scores show the model's improvement on a last-frame-copy baseline model.}
\label{fig:eval_t+n}
\end{figure}

Two key insights indicate that \textbf{PredNet is not a comprehensive emulation of hierarchical predictive coding}. Firstly, from the "mean E activation" plot in Figure \ref{fig:sliding_vis}, it is evident that the mean bottom-up error increases as one goes up towards higher layers. This behavior can be observed in all sample videos being visualized. This is in contrary to the expectations of predictive coding, which posits that the error decreases as we go up the hierarchy as parts of the incoming signal are iteratively `explained away.' Secondly, Lotter et al.~\cite{lotter16} demonstrate that models trained with $L_{0}$ loss perform better than $L_{all}$ loss on the KITTI data. We cross-check this on our dataset and get similar results. As shown in Figure \ref{fig:losses_comp}, models with $L_{0}$ loss perform better on all metrics. Training with $L_{0}$ loss implies that we only minimize the error $E_{0}$ on the lowest layer (see Figure~\ref{fig:prednet}) while in $L_{all}$ loss the model is trained to minimize the prediction errors in all the layers. Predictive coding suggests that each layer in the hierarchy minimizes the error signal iteratively. Therefore, an accurate implementation of predictive coding should improve results when trained with $L_{all}$ loss.
Furthermore, the visualization of mean activation of PredNet's states at different layers in Figure~\ref{fig:sliding_vis} shows that the states in the lowest layer are different from all the higher layers. This indicates that the model operates with two sub-modules when trained with the $L_{0}$ loss: the lowest layer  $R_{0}$ aims to generate realistic $(t+1)$ predictions, while the rest of the layers operate as one deep network that regresses $E_{0}$ to generate the context $R_{1}$. This is also indicated by the fact that the mean R activation for the lowest layer is higher and follows a different trajectory than the rest of the layers in Figure \ref{fig:sliding_vis}. 
We further evaluate PredNet by examining its ability to extrapolate and predict longer time steps into the future. As explained in Lotter et al. \cite{lotter16}, PredNet can be used to generate long-term predictions by simply feeding its predictions at time $t$ back in as input at the next time step $(t+1)$. This can be done iteratively for $n$ time steps to get a $(t+n)$ prediction into the future. We test the extrapolation capability of PredNet models that are trained only to perform $(t+1)$ predictions as well as PredNet models that are trained exclusively to perform $(t+n)$ predictions. As expected, the results are marginally better in the latter case as also demonstrated in Lotter et al. \cite{lotter16}. The extrapolated predictions of our best performing model are given in Figure~\ref{fig:t+n_extrap_different_timestep}. The extrapolation is started at different time points in the video as shown by the red marker in the figure. The following three observations can be made from the above experiments (1) After two-time steps, the model resorts to last-frame-copying. As already discussed, PredNet performs predictions by using the movement between consecutive input frames as an active cue. Therefore while extrapolating, when we feed the predictions back as input, the model gets a cue that the action has stopped and reverted to performing last-frame-copying. (2) The predictions get blurrier over time. This is because the minor blur added by the down-sampling units in the bottom-up and up-sampling units in the top-down accumulates exponentially with time. (3) From the metrics in Figure~\ref{fig:eval_t+n} we can infer that the models perform better if the extrapolation is started in the later stages of the video. This can be due to the fact that in our dataset, motion generally starts in the middle or towards the end of the video. In conclusion, the extrapolation experiments suggest that \textbf{the network design compels it to learn just short-term interpolations instead of building long-term predictions}.

\begin{figure}[t]
\begin{center}
   \includegraphics[width=4cm, height=7cm]{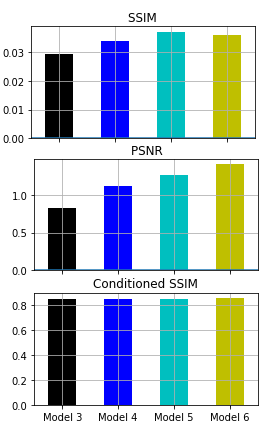}
\end{center}
   \caption{Comparison of performance with models encompassing 4, 5, 6 and 7 layers. SSIM and PRNS scores show the model's improvement on a last-frame-copy baseline model.}
\label{fig:layers_comp}
\end{figure}

Finally, we notice that the model delivers improved predictions only when the topmost layers have a full receptive field. Only this setting allows to predict object movements instead of just blurring local regions of motion. The receptive field can be increased either by using deeper layers or by increasing the kernel size of the convolutions or even by reducing the image size. We experimented with each of these and found that the prediction scores improve with the increased receptive field. We show the results of experimenting with a different number of layers in Figure \ref{fig:layers_comp} and it is apparent that the prediction quality improves with increasing depth.

\section{Label classification with PredNet+}\label{sec:lab_clas} \label{labelclassification}
In this section, we describe the second phase of experiments, where we test the architecture by modifying it to perform supervised label classification simultaneously with video prediction. For a comparison of the architectures of PredNet+ and the vanilla PredNet, see Figures \ref{fig:prednet+} and \ref{fig:prednet} respectively. The model design, the rationale behind the design and the results are discussed next. 

\subsection{PredNet+ Design} \label{prednet+model}
\begin{figure}[t]
\begin{center}
   \includegraphics[width=0.8\linewidth]{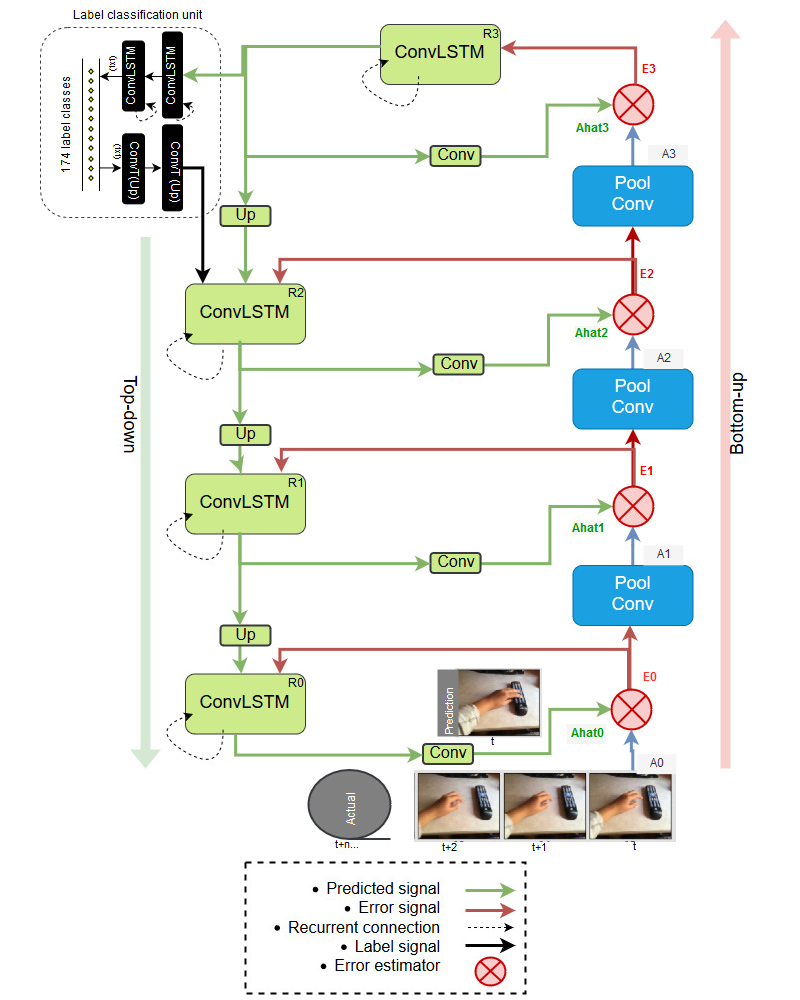}
\end{center}
   \caption{The proposed PredNet+ architecture with an additional classification pathway attached to the deepest prediction layer.}
\label{fig:prednet+}
\end{figure}

We modify the PredNet architecture such that it can perform video label classification along with next frame prediction and informally call this architecture PredNet+. The architecture is shown in Figure \ref{fig:prednet+}. As seen here, PredNet+ contains an additional label classification unit that is attached to the top-most representation layer. It consists of an encoder and a decoder part. As displayed in Figure \ref{fig:prednet+} the two ConvLSTM layers form the encoder which transforms the output of the representation unit $R_{3}$ into label class probabilities. The two transposed convolution layers make up the decoder that up-samples and transforms the label classes back to the imaging modality which is fed back into the top down as shown by the black arrow to $R_{2}$. 

%\begin{figure}[t]
%\begin{center}
%   \includegraphics[width=4cm, %height=3cm]{pics/weights.png}
%   \caption{Weighing of label predictions over %time.}
%%   \label{fig:weights}
%\end{center}
%\end{figure}

The label prediction unit makes predictions at each incoming frame, whose weighted sum is passed through a softmax function to get the final class probabilities for the video. As the model does not have enough context to make meaningful predictions at the beginning of the video, the weighing-over-time is done using an exponential function.

PredNet+ is designed such that the latent features at the top-most representation layer are shared between its two tasks. The future frame predictions are conditioned on the label predictions made by the label classification unit (shown in Figure~\ref{fig:prednet+} by the black arrow going into $R_2$). We hypothesized that this would improve the results on both sub-tasks as evident in many multi-task training scenarios \cite{Collobert2008} \cite{Girshick15}. Even though in our case, we attach the label classification unit to the top-most layer, this is not the only approach nor necessarily the best one as per predictive coding. We decided for this setup because the top-most layer in the model has both a full receptive field and access to previous states.

In summary, the label classification unit and the prediction units in PredNet+ are expected to work in tandem in a multitask learning set-up and form a synergy. However, this is not what we observe in our results.

\subsection{Results}
Table \ref{tab:results_acc} shows our best classification accuracy in comparison to the baseline model scores of Goyal et al.~\cite{goyal17} and the current state-of-the-art results by Mahdisoltani et al. on the Something-something dataset\cite{farzaneh18}.\linebreak
We test the PredNet+ architecture on our best 4 layer model, 5 layer model and 6 layer model from Table \ref{tab:models}. Furthermore, we test the following minor variations of PredNet+ to further evaluate the model architecture: First, we remove the recurrent memory in the label classification unit by replacing the ConvLSTM with Convolution layers. Next, we extend the label classification loss function such that the model is rewarded for predicting at least the correct verb in the label. For example, if the correct label is \textit{Pretending to \underline{put} something behind something}, the model is penalized twice as much if it predicts \textit{\underline{Showing} something to the camera} than if it predicts \textit{\underline{Putting} something behind something}, which has the same verb as the correct label. Surprisingly, the classification results do not change at all (${\pm0.6}{\%}$) for any of these model variations. This suggests that the features from the top-most representation unit do not have any more information. \linebreak

\begin{figure}[t]
\begin{center}
   \includegraphics[width=4cm, height=7cm]{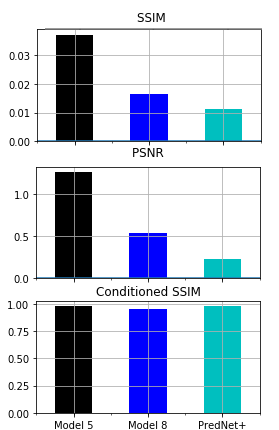}
\end{center}
   \caption{Comparison of the prediction quality of PredNet+ with equivalent PredNet models. SSIM and PRNS scores show the model's improvement on a last-frame-copy baseline model.} 
\label{fig:models_comp}
\end{figure}

\begin{table}
\begin{center}
\begin{tabular}{|l|c|c|}
\hline
Model & Top-1 & Top-5 \\
\hline\hline
Baseline \cite{goyal17} & 11.5 & 30.0 \\
Ours & 28.2 & 57.0 \\
Mahdisoltani et al. \cite{farzaneh18} & \textbf{51.38} & - \\
\hline
\end{tabular}
\vspace*{5mm}
\caption{\label{tab:results_acc}Classification accuracy on the Something-something dataset with 174 label categories.}
\end{center}

\end{table}
The label classification scores suggest that PredNet+ is a long way from the state-of-the-art architectures. Furthermore, the future frame prediction of PredNet+ degrade in comparison to its equivalent vanilla PredNet models: Model 5 ($L_0$ loss) and Model 8 ($L_{all}$ loss). The metrics in Figure \ref{fig:models_comp} and the visualization of predictions point this out. To further analyze this, we experiment with different loss weights for the two tasks. This allows us to control the relative importance of each task for the model during training. We find that the model's future prediction quality degrades when the label classification task is given increased importance, suggesting that the multi-task constraint leads to worse future frame predictions. 

\subsection{Representation learning with top-down conditioning and synthetic data} 
\begin{figure}[t]
\centering
   \begin{subfigure}(a) PredNet
   \includegraphics[width=1.\linewidth]{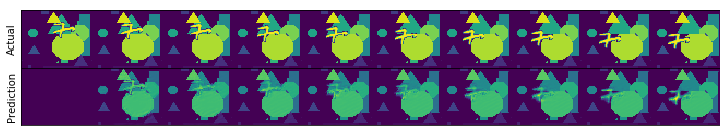}
    \end{subfigure}

    \begin{subfigure}(b) PredNet+
   \includegraphics[width=1.\linewidth]{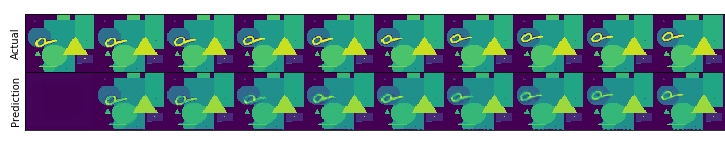}
    \end{subfigure}
\caption{Model predictions on moving MNIST. Shown are examples for the original model and a model with additional classification pathway.}
\label{fig:mnist}
\end{figure}

In order to further evaluate our hypothesis that conditioning the top-down predictions on class labels of the video improves model accuracy, we evaluate PredNet+ on a synthetic dataset. We employ a moving MNIST dataset with a static background consisting of randomly generated overlapping geometric shapes, and a single hand-written digit moving in one of eight directions. Each frame was annotated with a label representing the future direction of the digit's movement. Samples of the dataset are given in the upper rows of Figure \ref{fig:mnist}.
We test if adding semantic top-down information helps increasing the prediction score of the network. We thus keep track of spatiotemporal prediction performance, while using the movement label classification as an auxiliary task. The generated predictions generally showed lower confidence in the moving part of the input frames, especially in the first frames, as seen in the lower row of Figure \ref{fig:mnist} (a). Predictions made by the model with additional label classification pathway are presented in Figure \ref{fig:mnist} (b). 

The resulting scores generated by a previous-frame-copy model, plain PredNet and PredNet+ are displayed in Table \ref{tab:mnist_res}. The multi-task learning with movement direction classification improves the MAE score and leads to sharper predictions in the non-stationary parts of the input images. This indicates that conditioning the top-down predictions with semantic information can improve model performance, especially when the additional information can be related directly to predicted features in the input space. 

\begin{table}
\begin{center}
\begin{tabular}{|l|c|c|}
\hline
Model & MAE score \\
\hline\hline
Previous-frame-copy & 8e-050\\
PredNet & 7.6e-05 \\
PredNet+ & 7.3e-05 \\
\hline
\end{tabular}
\vspace*{5mm}
\end{center}
\caption{\label{tab:mnist_res} Comparison of the predictive performance of the original and multi-task PredNet model. Scores are given for next-frame prediction on the moving MNIST dataset. }
\end{table}

\section{Conclusion and future work}\label{sec:conc}
We have evaluated PredNet \cite{lotter16} on a challenging action classification dataset in two phases.

In the first phase of our work, we investigate PredNet and derive the following insights: (1) PredNet does not completely follow the principles of the predictive coding framework. (2) It can perform only short-term next frame interpolations, rather than long-term video predictions. This has been further confirmed by the extrapolation experiments. (3) The representation units are unable to learn multi-modal distributions and produce blurry predictions. (4) The models' learning ability is sensitive to the continuity of motion and the FPS rate of the videos. 

In the second phase, we test PredNet's ability to learn useful latent features to perform label classification. We use the features from the highest representation layer and find that this is not adequate for the task at hand, namely, the prediction of a complex action classification dataset. We achieve a classification accuracy of 28.2\% in comparison to current state-of-the-art of 51.38\% \cite{farzaneh18} and the prediction accuracy also under-performs the vanilla PredNet. In a further step, we experiment on a synthetic dataset and show that that top-down conditioning can improve the prediction scores. 

Our results lead to several suggestions for improved models: Firstly, the network should be trainable with $L_{all}$ loss. This can be done by designing error estimators that are local to each layer. Secondly, the network should be redesigned such that it is encouraged to perform long-term predictions rather than just frame-to-frame interpolation. One way to do this is to have additional layers higher in the hierarchy, that make predictions at different temporal scales. Additionally, PredNet's performance metrics show high variance while PredNet+ is easily susceptible to over-fitting. These points signal the need for including regularization techniques and model averaging methods like dropout within the architecture. Finally, the representation units should learn multi-modal probability distributions, from which predictions can be sampled. This could be addressed, for example, with probabilistic representations in some or all layers.
\bibliographystyle{ACM-Reference-Format}
\balance
\bibliography{ourbib}

\appendix
\clearpage

%\appendix
%\begin{figure*}[h]
%\begin{center}
%\textbf{Appendix A: All results}\par\medskip
%    \label{app:all_results}
%   \includegraphics[width=\linewidth,height=.9\textheig%ht,keepaspectratio]{pics/all_metrics.png}
%  \caption{Results of all the metrics from Section %\ref{metrics} for all the 10 models listed in Table %\ref{tab:models}.}
%  \label{figa1}
%  \end{center}
%\end{figure*}

%\label{app:vis}

%\begin{figure*}[h]
%\begin{center}
%\textbf{Appendix B: Examples of full visualization}\par\medskip
%    \label{app:all_results}
%  \includegraphics[width=\textwidth]{pics/best_putting_full.png}
%  \caption{Example of full visualization.}
%  \label{figa1}
%%\end{center}
%\end{figure*}

%\begin{figure*}[h]
%\begin{center}
%  \includegraphics[width=\textwidth]{pics/best_rolling_%full.png}
%  \caption{Example of full visualization.}
%%  \end{center}
%  \end{figure*}

%%\begin{figure*}[h]
%\begin{center}
%\textbf{Appendix B: Example of full visualization on %extrapolation.}\par\medskip
%    \label{app:extrap}
%  \includegraphics[width=\textwidth]{pics/turning_extra%p.png}
 % \caption{Extrapolation results for Model 7 trained at 3t/4 steps.}
  %\end{center}
%\end{figure*} 

\end{document}